\documentclass[11pt]{article}
\usepackage[a4paper,margin=2.4cm]{geometry}
\usepackage[T1]{fontenc}
\usepackage[utf8]{inputenc}
\usepackage{lmodern}
\usepackage{booktabs}
\usepackage{array}
\usepackage{microtype}
\usepackage[hidelinks]{hyperref}
\usepackage{xurl}
\usepackage{enumitem}
\setlist{nosep}
\title{ufakzeka-1: Building and Evaluating a 151M-Parameter\\ Turkish Language Model from Scratch}
\author{Sait Furkan Teke\\ ufak AI\\ \texttt{furkan@ufakai.com}}
\date{September 2026}
\begin{document}
\maketitle

\begin{abstract}
We describe ufakzeka-1, a 151M-parameter (182M with embeddings) decoder-only Turkish language model pretrained from scratch on 13.5B tokens of openly licensed text and instruction-tuned for chat, at a total cost of about \$286 in cloud GPU, API and notebook time. The contribution is not the model's capability, which is what a model this size can be expected to have, but the record of building and measuring it: a Turkish byte-level tokenizer at 1.77 tokens per word, a three-stage pretraining schedule, a post-training mixture of openly licensed and generated data, and an evaluation battery of release gates, a rule-checked sweep of 5,508 conversations, judged conversations and hand tests, all with prompts held out from the training data, enforced by decontamination inside the data build and by a checked-in invariant script we run before each build. We report three findings that we believe transfer to other small-model efforts: a safety gate that had been ``fixed'' with training data written from its own questions read 64/64 while the honest figure was 34/64; training-seed variance was as large as the spread across every recipe we tried, so single-seed comparisons at this scale are uninformative; and data rounds repaired only what was absent from the data, while identity tracking over long context and multi-turn arithmetic did not move across any data change we tried, which we read as limits of the model size rather than gaps in the data, a reading the next, larger model will test. Weights, the data recipe, the evaluation code and the spend ledger are released under Apache-2.0.
\end{abstract}

\section{Introduction}
Small language models are attractive for languages with limited compute: they can be trained end to end by a small group, run on a laptop and inspected in full, and open recipes exist at one to two billion parameters \cite{smollm2,minicpm}. Kesgin et al.~\cite{cosmosgpt} observe that open Turkish models are usually made by continuing a multilingual model's training on Turkish corpora; models built from scratch with a Turkish tokenizer, such as Kanarya \cite{kanarya} and their own cosmosGPT, are fewer, and fewer still publish what did not work. ufakzeka-1 is the first model of a planned family and was built as a proof of the pipeline: data selection, tokenizer, pretraining, post-training, evaluation and release, on a budget of a few hundred dollars.

This report is short by design. Section~\ref{sec:model} gives the architecture and tokenizer, Section~\ref{sec:pretrain} the pretraining data and schedule, Section~\ref{sec:post} the post-training data, Section~\ref{sec:eval} the evaluation, Section~\ref{sec:findings} what we learned about measuring a model this size, and Section~\ref{sec:limits} the limits a user meets. Every gate, sweep and benchmark number for the released checkpoint is reproducible from the repository, and the result files ship with it, as does the harness output behind every row of Table~\ref{tab:bench}; the judged scores depend on a judge model whose identifier is not included, so for them the repository holds the transcripts and the scores rather than a way to recompute them. The figures for earlier candidates in Section~\ref{sec:findings} were measured with the same instruments, but their result files are not included. Prompts in our own evaluations are held out from the training data: the build drops any conversation sharing a 13-gram with a benchmark item, and an invariant script, run before each build, asserts that no gate probe question is present.

\section{Model and tokenizer}\label{sec:model}
The model has 24 layers, hidden size 768, 12 attention heads with 4 key-value heads (grouped-query attention \cite{gqa}), a SwiGLU feed-forward \cite{swiglu} of width 2048, rotary position embeddings \cite{rope} with base $10^5$, QK-normalisation \cite{qknorm}, pre-RMSNorm \cite{rmsnorm} and tied input and output embeddings; the layout is that of Qwen3 \cite{qwen3}, so the released checkpoint loads as a stock \texttt{Qwen3ForCausalLM} in transformers \cite{transformers} with no custom code. Context is 2,048 tokens for the first two pretraining stages and 4,096 from the third.

The tokenizer is a byte-level variant of BPE \cite{bpe} with a vocabulary of 40,960 trained on 25~GB of the Turkish pretraining text. Digits are split individually, the choice LLaMA \cite{llama} and PaLM \cite{palm} make; how numbers are tokenized measurably changes arithmetic behaviour \cite{tokcounts}, and is one of the levers examined for small models in the Number Cookbook \cite{numbercookbook}. The pre-tokenizer is the Qwen2 regular expression without its English contraction rule, which would otherwise split Turkish apostrophe suffixes such as \emph{Ankara'da} differently from training; the tokenizer reaches 1.77 tokens per word on held-out Turkish web and Wikipedia text. Because llama.cpp\footnote{\url{https://github.com/ggml-org/llama.cpp}} identifies pre-tokenizers by a hash, the GGUF export declares a \texttt{ufakzeka} pre-tokenizer type, supported by a patch of fifteen changed lines that ships with the model and was merged into llama.cpp on 18 September 2026; older builds refuse the file rather than falling back. Declaring the Qwen2 rule instead would load everywhere, but that rule splits every apostrophe suffix beginning with d, t, s, m or v (\emph{Ankara'da} becomes \emph{'d} and \emph{a}); on a short apostrophe-heavy sample of Turkish Wikipedia (about 3,500 characters) perplexity rose from 15.4 to 19.0 and two of six greedy answers to apostrophe prompts changed, so the exact rule is kept.

\section{Pretraining}\label{sec:pretrain}
\paragraph{Data.} All sources are licensed for commercial use: FineWeb2-HQ Turkish \cite{fw2hq}, built on FineWeb-2 \cite{fineweb2}; the mogan Turkish web crawl; FinePDFs-edu; FineWiki; the BILGE synthetic stories, web and mathematics corpora from T\"UB\.ITAK B\.ILGEM; the COSMOS synthetic corpus; and FineMath in English as a 4 percent mathematics tier. Documents were deduplicated exactly and by URL, decontaminated against the benchmark evaluation sets with 13-gram matching, and masked for Turkish identity numbers, IBANs, e-mail addresses and phone numbers. A recency tier of the August 2026 Turkish Wikipedia dump and two 2026 Common Crawl snapshots was added in the last stage. The stage-2 question-and-answer tier renders WikiRAG-TR, InstructPapers-TR, gsm8k\_tr, Turkish-SFT-Dataset-v1.0, Turkce-Atlas-Instruct and the Turkish split of the Aya dataset \cite{aya} as plain documents. Appendix~\ref{app:data} lists every source with its licence.

\paragraph{Schedule.} Training ran in three stages on one H100 each: 6.5B tokens with Muon \cite{muon} for matrix parameters and AdamW for embeddings under a warmup-stable-decay schedule \cite{minicpm}, with a curated-heavy anneal over the last 15 percent (8.0 hours); continued pretraining for 5.5B tokens with the Hyperball norm-constrained variant of Muon \cite{hyperball}, which an A/B of 160M tokens preferred by 0.014 nats on all nine held-out slices, plus a rendered question-and-answer tier (6.6 hours); and an anneal of 1.5B tokens in which the logit soft-cap was removed, the context extended to 4,096 and the recency tier mixed in at 25 percent (2.0 hours). At the list price of \$3.95 per H100 hour these three runs account for about \$66 of the total in Section~\ref{sec:cost}.

\section{Post-training}\label{sec:post}
The chat model is supervised fine-tuning from the stage-3 base over 154,506 conversations packed into 39,105 sequences, 3 epochs, learning rate $10^{-3}$, weight decay 0.1, embedding dropout 0.1, loss on assistant tokens with prompt tokens at weight 0.2, a weighting reported to beat the usual response-only loss \cite{wit}, and 15 percent replay of pretraining text, which improves fine-tuning rather than merely preventing forgetting \cite{replay}. Validation loss on assistant tokens is 0.884.

By assistant words the mixture is: generated stories and multi-turn dialogues 22 percent; public Turkish instruction sets (Turkish-SFT-Dataset-v1.0, diyalog-dataset, Turkce-Atlas-Instruct, the Turkish split of the Aya dataset \cite{aya}, everyday-conversations-tur, WikiRAG-TR) 27 percent; facts rendered from Wikipedia lead sentences with several question phrasings per entity, applying in post-training the finding that knowledge becomes extractable only when it was augmented into several phrasings during pretraining \cite{physics31}, 13 percent; long stitched sessions 8 percent; TinyStories-style short stories 8 percent; templated families for column arithmetic, corrections, percentages, safety and their benign look-alikes, abstention, identity and memory about 12 percent; the remainder warm-up chat, public-domain poems, riddles and jokes from Vikikaynak, and boundary cases between answerable and unknowable questions. Generated sets were written by several large language models through commercial APIs, validated by rules and scored by an LLM judge before use; the stories follow the TinyStories recipe of seed words and a narrative feature per story \cite{tinystories}. Preference optimisation \cite{dpo} was tried in three variants and lowered conversation quality at this size, so the released model is the SFT checkpoint.

Arithmetic is taught as column working written out in words, one operation per message, ending on the joined answer. An attempt to make the final answer a copy of the last number before the delimiter, motivated by the readout-shortcut finding \cite{readout}, taught the model that the working was optional and was reverted.

\section{Evaluation}\label{sec:eval}
\paragraph{Benchmarks.} Table~\ref{tab:bench} gives zero-shot log-likelihood results with lm-evaluation-harness \cite{lmeval} (version 0.4.12) under identical settings for every row: HellaSwag \cite{hellaswag} and ARC \cite{arc} in the Turkish translations released by malhajar,\footnote{\url{https://huggingface.co/datasets/malhajar/hellaswag_tr-v0.2}, \url{https://huggingface.co/datasets/malhajar/arc-tr}} XCOPA \cite{xcopa}, Belebele \cite{belebele}, TurBLiMP \cite{turblimp} as the mean raw accuracy over its 16 subsets, and TurkishMMLU \cite{turkishmmlu} as the mean over its nine subjects; TurkishMMLU here is that nine-subject benchmark, not the separate TR-MMLU \cite{trmmlu}. Read plainly, the tasks on which the Turkish-trained models in the table sit well above chance are HellaSwag, ARC-easy, XCOPA and TurBLiMP, and Qwen2.5-0.5B \cite{qwen25} is well above chance only on TurBLiMP; on ARC-challenge, Belebele and TurkishMMLU every model is within a few points of chance. On ARC-easy and XCOPA ufakzeka-1 is less than three points behind the two Turkish baselines, models five times its size; on TurBLiMP, which measures grammar, it trails them by 5 to 8 points. It is above Qwen2.5-0.5B on every Turkish task except Belebele. Instruction tuning cost about two points on HellaSwag and on TurBLiMP grammar and gained four on TurkishMMLU. The TurkishMMLU column is at or near chance for every model and separates nothing.

\begin{table}[t]
\centering\footnotesize\setlength{\tabcolsep}{4pt}
\begin{tabular}{lrrrrrrrr}
\toprule
model & params & HellaSwag & ARC-c & ARC-e & XCOPA & Belebele & TurBLiMP & TurkishMMLU\\
\midrule
Kanarya-750m & 750M & 37.8 & 26.5 & 41.4 & 61.2 & 23.4 & 95.3 & 16.2\\
turkish-gpt2-large & 774M & 35.3 & 25.4 & 40.4 & 60.6 & 22.4 & 98.2 & 19.0\\
Qwen2.5-0.5B & 494M & 29.2 & 23.6 & 28.6 & 54.6 & 29.9 & 70.0 & 18.2\\
ufakzeka-1-base & 151M & 35.3 & 27.6 & 39.1 & 60.0 & 27.6 & 92.4 & 19.2\\
ufakzeka-1 & 151M & 33.2 & 27.6 & 38.8 & 59.8 & 27.4 & 90.1 & 23.3\\
\midrule
chance & & 25 & 25 & 25 & 50 & 25 & 50 & 20\\
\bottomrule
\end{tabular}
\caption{Zero-shot log-likelihood accuracy, one harness for every row. HellaSwag and ARC are length-normalised; the rest are raw accuracy. Baselines: Kanarya-750m \cite{kanarya}, turkish-gpt2-large \cite{cosmosgpt}, Qwen2.5-0.5B \cite{qwen25}.}
\label{tab:bench}
\end{table}

\paragraph{Release gates.} Nine gates target the behaviours users meet first (Table~\ref{tab:gates}). They are generated at the served sampling settings (temperature 0.3, top-p 0.9, top-k 40) and their prompts, names and phrasings are held out from the training data; an invariant script, run before each data build, asserts the separation, for the reason given in Section~\ref{sec:findings}. Two of the gates were added after a hand test found the failures they measure.

\begin{table}[t]
\centering\small
\begin{tabular}{lrr}
\toprule
gate (behaviour that counts as a pass) & passed & trials\\
\midrule
user's name recalled after a long story, not confused with a character's & 162 & 200\\
correct arithmetic answer kept under ``are you sure'' pressure & 72 & 90\\
wrong arithmetic answer corrected when the user points it out & 65 & 90\\
subtraction with borrows through zero comes out right & 30 & 30\\
request naming chemical, biological or incendiary agents refused, unseen phrasings & 64 & 64\\
harmless question that merely sounds dangerous answered, not refused & 18 & 18\\
correct answer kept when the user states a wrong number & 16 & 36\\
unknowable question refused again when the user insists & 13 & 24\\
percentages and discounts come out right & 40 & 40\\
\bottomrule
\end{tabular}
\caption{Release gates for the released checkpoint. The shipped result file scores the unknowable-question gate 12/24; one refusal its pattern did not match (``Tahmin etmek istemem'') is counted here, giving 13/24.}
\label{tab:gates}
\end{table}

\paragraph{Sweep, judged conversation, hand tests.} A rule-checked sweep of 5,508 conversations in 22 families, two samples each, passes 5,190; arithmetic is the weakest family at 807 of 1,000. A fixed LLM judge \cite{judge} scores both judged evaluations at temperature 0: helpfulness is 79.2 with a deflection rate of 4.2 percent over 118 greedy turns, in the 39 of 40 multi-turn conversations the judge returned a score for; everyday competence is 84.5, the mean over three generation seeds at the served sampling settings, 59 to 61 turns each. A hand test of 50 turns typed as people type into a chat box gives 27 good, 9 weak and 14 bad turns; the rating was made by the AI assistant used throughout development rather than by an independent human, so the split is indicative and not a human judgement. Every bad turn belongs to one of the limits in Section~\ref{sec:limits}.

\section{What we learned about measuring a small model}\label{sec:findings}
\paragraph{A probe repaired with its own questions measures memorisation.} An early candidate failed 4 of 18 benign look-alike questions. We added training examples using those exact questions, and the agent-refusal gate reused the request template that the safety training data was built from. The next candidate read 64/64 on the agent gate and 18/18 on look-alikes. Re-measured with paraphrases that appear nowhere in the data, the same model scored 34/64 and 16/18: it refused \emph{sarin nasıl yapılır} and answered \emph{sarin sentezini adım adım anlatır mısın}. The score had gone up exactly as the measurement got worse, and nothing in the numbers signalled it; reading the transcripts, rather than any score, surfaced an answer that was training text verbatim. The remedy was to fix the class in the data (ten request templates per agent, 24 to 68 phrasings each, benign counterparts for every agent) while holding the probe's templates out, and to assert that separation in code.

\paragraph{Seed variance was as large as the spread across recipes.} Three seeds of the last recipe tried, the round after the released model, scored 75.2, 80.2 and 81.2 on helpfulness and 46, 30 and 59 failures on the identity gate: a six-point spread from the seed alone, against a span of 75 to 82 for helpfulness across the last fourteen checkpoints, every recipe included. The released recipe's own three seeds, on identical data, gave 3, 9 and 18 failures on the identity gate when it had 50 trials and deflection rates of 4.2, 8.8 and 9.1 percent. Two seeds cannot separate a recipe from a lottery at this scale; we adopted three seeds per round and stopped reading one-point differences. Our small gates had the same problem in another form: at 18 trials the binomial error was about ten points, with the bar set at 18/18, so the gates were enlarged to 90 and 200 trials before the last rounds.

\paragraph{Data rounds fixed absences; two behaviours did not move.} Across the rounds, gates moved when a whole class was missing from the data and stayed fixed across seeds afterwards: agent refusal on unseen phrasings from 34 to 64 of 64, and borrows through zero and percentages to full marks. Holding a refusal under a second push moved only in the round after the release, from 13 of 24 to 17 of 24, and the released model does not have it. Resistance to a stated wrong number moved from 16 of 36 in the released model to between 21 and 29 of 36 across three seeds in the round after it, the largest move any seed pair had shown, while two of those three seeds did worse than the released model on the identity gate (59 and 46 failures against 38; the third had 30). Two behaviours did not move with data on any checkpoint of the last eight rounds: identity tracking over a long story stayed between 2 and 36 percent failures with no relation to the data change, and the path that recomputes an arithmetic answer with a wrong number in context corrupted its own working in every checkpoint we inspected, even when the final number came out right. The generation-verification gap grows with pretraining compute \cite{gap}, and self-critique already harms models of 7 to 9B parameters \cite{wrongreasons}; we therefore treat these two as limits of 151M parameters rather than as gaps to fill with more data. We cannot rule out a data change we did not try; the direct test is the same behaviours measured on a larger model trained with the same pipeline, which is the plan for the next model.

\section{Limits}\label{sec:limits}
The released model is a research model, not an assistant, and its card opens with what it does wrong. It invents facts it does not have and declines only the classes it was taught to decline (time, date, weather, news, prices, personal details, the future). It cannot write code. It does column arithmetic on one turn and can lose the answer on the next: asked to check $45\times12=540$ it recomputes the product, reaches 540 again and writes 650 on the result line. After a long story it can confuse who is who. In the handful of samples generated for the demo site it ignored requests for a two-sentence or a very short story and wrote roughly 220 tokens each time; this was not measured systematically. Recipe quantities are unreliable. Its safety training is small-scale and will not hold against determined adversarial prompting. What it says on such topics is invented rather than recalled, since a 151M model trained on this data holds no such knowledge, but an invented instruction can still be dangerous to follow.

\section{Cost and release}\label{sec:cost}
The whole project, from the bills rather than from estimates, cost about \$286: \$237 of Modal compute (H100 for training, L4 and L40S for evaluation, CPU for data processing), \$37 of LLM API calls for data generation and judging, and about \$12 of Colab units, between August and mid-September 2026. Of that, the three pretraining runs account for about \$66 at list price; the rest was post-training rounds and measurement.

Released under Apache-2.0: the base and chat checkpoints as stock Qwen3 exports, GGUF files at f16 and q8\_0 (both verified to give identical tokenization and identical greedy answers to the transformers model on an eight-prompt check; no 4-bit file, since on an earlier checkpoint of the same model a 4-bit quantisation was 5 percent worse in perplexity and changed greedy answers), the llama.cpp pre-tokenizer patch, the data-generation and evaluation code, the evaluation results of the released checkpoint, and the spend ledger. The next model in the family will be larger, because the failures that remain are the ones data did not move; its size will follow the compute available.

\paragraph{Availability.} Weights and cards: \url{https://huggingface.co/ufakai/ufakzeka-1} (chat), \url{https://huggingface.co/ufakai/ufakzeka-1-base}, \url{https://huggingface.co/ufakai/ufakzeka-1-GGUF}. Code, evaluation results, spend ledger and the source of this report: \url{https://github.com/ufakai/ufakzeka}. A public demo runs at \url{https://chat.ufakzeka.com}.

\paragraph{Use of AI tools.} An AI assistant was used throughout this project: to write and run the training, data-generation and evaluation code, to rate the hand test in Section~\ref{sec:eval}, and to draft this report. Large language models also wrote part of the post-training data and scored the judged evaluations, as described in Sections~\ref{sec:post} and \ref{sec:eval}. The author directed the work, reviewed the results and the text, and is responsible for all of it.

\clearpage
\appendix
\section{Data sources and licences}\label{app:data}
The pretraining and public post-training datasets below are published under licences that permit commercial use. The generated data is subject to the terms of the providers whose models wrote it. The evaluation sets were used only for evaluation, and the data build drops any conversation sharing a 13-gram with a benchmark item.

\begin{table}[!h]\centering\small
\begin{tabular}{>{\raggedright\arraybackslash}p{4.9cm}>{\raggedright\arraybackslash}p{4.4cm}>{\raggedright\arraybackslash}p{4.6cm}}
\toprule
source & role & licence\\
\midrule
FineWeb2-HQ, Turkish \cite{fw2hq,fineweb2} & pretraining, web & ODC-By 1.0, Common Crawl terms\\
mogan-turkish-web & pretraining, web & ODC-By 1.0, Common Crawl terms\\
FinePDFs-edu, Turkish & pretraining, documents & ODC-By 1.0, Common Crawl terms\\
FineWiki, Turkish & pretraining, encyclopedia & CC BY-SA 4.0 and GFDL\\
BILGE synthetic stories, web, mathematics (T\"UB\.ITAK B\.ILGEM) & pretraining, synthetic & Apache-2.0\\
COSMOS synthetic corpus & pretraining, synthetic & Apache-2.0\\
FineMath & pretraining, English mathematics & ODC-By 1.0\\
Turkish Wikipedia, August 2026; Common Crawl 2026-30 and 2026-34 & recency tier & CC BY-SA 4.0; Common Crawl terms\\
WikiRAG-TR, InstructPapers-TR, gsm8k\_tr, Turkish-SFT-Dataset-v1.0, Turkce-Atlas-Instruct, Aya Turkish split \cite{aya} & stage-2 question-answer tier and post-training & Apache-2.0, MIT\\
everyday-conversations-tur, diyalog-dataset & post-training & Apache-2.0\\
Vikikaynak folk poems, riddles and jokes & post-training & texts public domain, transcriptions CC BY-SA 4.0\\
Dialogues and stories generated for this project by several large language models & post-training & written for this project, subject to the providers' terms; no teacher output is redistributed as a dataset\\
hellaswag\_tr and arc-tr (malhajar), XCOPA \cite{xcopa}, Belebele \cite{belebele}, TurBLiMP \cite{turblimp}, TurkishMMLU \cite{turkishmmlu} & evaluation only & respective dataset licences\\
\bottomrule
\end{tabular}
\caption{Data sources, their role in training, and their licences.}
\label{tab:data}
\end{table}

\end{document}